\documentclass{article}
\usepackage{ijcai23}

\usepackage{times}
\usepackage{soul}
\usepackage{url}
\usepackage[hidelinks]{hyperref}
\usepackage[utf8]{inputenc}
\usepackage[small]{caption}
\usepackage{graphicx}
\usepackage{amsmath}
\usepackage{amsthm}
\usepackage{booktabs}
\usepackage{algorithm}
\usepackage{algorithmic}
\usepackage[switch]{lineno}

\usepackage{cleveref}
\usepackage[subrefformat=parens,labelformat=parens]{subcaption}
\usepackage{mathalpha}
\usepackage{wrapfig}
\usepackage{multirow}
\usepackage{enumitem}
\usepackage{microtype} 
\usepackage{amsfonts}
\usepackage{amssymb}  
\usepackage[normalem]{ulem}
\useunder{\uline}{\ul}{} 

\usepackage{comment}

\usepackage{xcolor}

\title{Reducing Communication for Split Learning by Randomized Top-$k$ Sparsification}

\author{
Fei Zheng$^1$
\and
Chaochao Chen$^1$\thanks{Corresponding author.} \and
Lingjuan Lyu$^{2}$ \and
Binhui Yao$^3$ 
\affiliations
$^1$Zhejiang University\\
$^2$Sony AI\\
$^3$Midea Group
\emails
\{zfscgy2, zjuccc\}@zju.edu.cn,
lingjuan.lv@sony.com,
tony.yao@midea.com,
}

\begin{document}

\maketitle
\setlength{\textfloatsep}{3pt}
\setlength{\floatsep}{3pt}
\setlength{\intextsep}{10pt}

\begin{abstract}
Split learning is a simple solution for Vertical Federated Learning (VFL), which has drawn substantial attention in both research and application due to its simplicity and efficiency.
However, communication efficiency is still a crucial issue for split learning.
In this paper, we investigate multiple communication reduction methods for split learning, including cut layer size reduction, top-$k$ sparsification, quantization, and L1 regularization.
Through analysis of the cut layer size reduction and top-$k$ sparsification, we further propose randomized top-$k$ sparsification, to make the model generalize and converge better.
This is done by selecting top-$k$ elements with a large probability while also having a small probability to select non-top-$k$ elements.
Empirical results show that compared with other communication-reduction methods, our proposed randomized top-$k$ sparsification achieves a better model performance under the same compression level.

\end{abstract}

\section{Introduction}
In recent years, the protection of data privacy has become an important issue in machine learning.
To date, many kinds of solutions have been proposed to solve the data privacy issue. 
%
Aside from cryptographic methods~\cite{mohasse2018aby3,wagh2019securenn,ccc2021homommorphic}, Federated Learning (FL)~\cite{litian2020federated_learning} 
and Split Learning (SL)~\cite{vepakomma2018split} are two promising methods for privacy-preserving machine learning. 
While typical federated learning (also known as Horizontal Federated Learning, HFL) mainly focuses on horizontally distributed data, split learning is a simple and effective solution for vertically distributed data.
However, although split learning achieves better efficiency than cryptographic methods~\cite{zhoujun2020vertical_gnn}, the communication efficiency is still an important issue.
%
In this paper, we focus on reducing the communication of split learning.

\begin{figure}[t]
    \centering
    \includegraphics[width=1\linewidth]{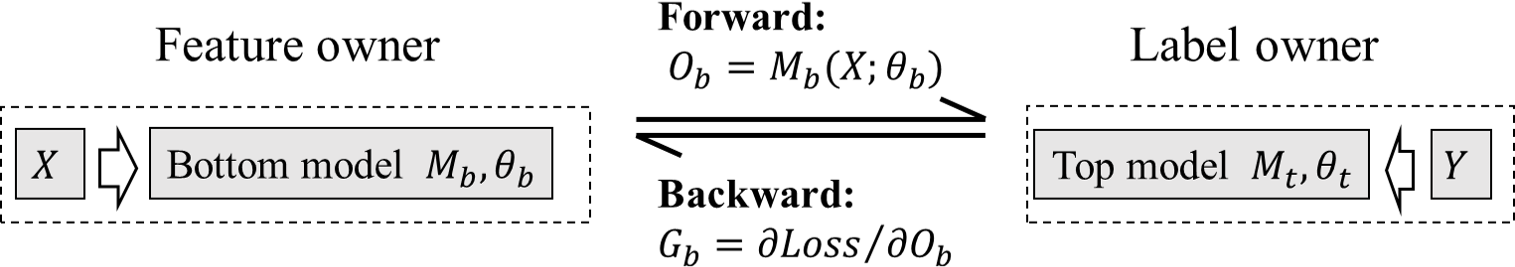}
    \caption{Overview of split learning.}
    \label{fig:split-learning-overview}
\end{figure}

%
The basic idea of split learning is to divide the model into several parts, and each part is computed by a different participant (party).
Since split learning only requires sharing the intermediate outputs or gradients, instead of the original input and label, it is considered somewhat privacy-preserving.
We present the overview of a typical split learning model in \Cref{fig:split-learning-overview}, 
where we consider a two-party case --- input feature and label are held by two parties, i.e., feature owner and label owner. 
To train a model without directly revealing participants' data, the model is split into a \textit{bottom model} and a \textit{top model}, held by the feature owner and the label owner, respectively. 
The last layer (output) of the bottom model is called the 
\textit{cut layer}.
We simply describe the training procedure of split learning below:
\begin{itemize}[leftmargin=20pt]
    \item Forward pass: the feature owner first feeds the sample features $X$ to the bottom model $M_b$, then gets the intermediate output $O_b = M_b(X, \theta_b)$, where $\theta_b$ is the bottom model's weight.
    The label owner fetches $O_b$, feeds it to the top model $M_t$, and then gets the final output $\hat Y = M_t(O_b;\theta_t)$.
    \item Backward pass: the label owner calculates the loss function $L(\hat Y, Y)$, then computes the gradients $\partial L(\hat Y, Y)/\partial \theta_t$ and $G_b = \partial L(\hat Y, Y)/\partial O_b$. 
    The former gradient is used to update the top model, and the latter is sent to the feature owner.
    The feature owner computes the jacobian matrix $\partial O_b/\partial \theta_b$. 
    Then the gradient with respect to the bottom model's parameters can be computed as $\left(\partial O_b/\partial \theta_b\right)^TG_b$, which is used to update the bottom model.
\end{itemize}

However, despite its simplicity, the communication efficiency of split learning is still an important issue.
During each iteration, the participants have to exchange intermediate results of the data batch,
and the number of iterations and the size of intermediate results are usually large.
For example, consider training ResNet-20~\cite{hekaiming2016resnet} with the cut layer size equal to $32\times32\times32$ and batch size equal to 32. It takes 2 (forward \& backward) $\times$ 4 (bytes for float value) $\times$ 32 (batch size) $\times 32^3$ (size of a single sample's intermediate result) bytes = $8$ Mib traffic to finish a single iteration of inference, which is a huge communication cost, especially for mobile devices.

For federated learning, existing work has investigated different methods to reduce communication, including 
increasing local epochs~\cite{mcmahan2017fedavg},
accelerating convergence~\cite{Karimireddy2020scaffold,wanghongyi2020fedma,wangjianyu2020fednova},
and compressing local model updates in various ways~\cite{aji2017gradient_drop,stich2018sparse_memory,sattler2019sparse_binary,sattler2019sparse_ternary,wanghongyi2018atomo}.
For split learning, however, studies on communication reduction are relatively few.
Castiglia et al. \shortcite{timothy2022compressed-vfl} proved the convergence of compressing split learning by top-$k$ sparsification or quantization.
Other methods such as asynchronous training, quantization, and using an autoencoder as a compressor are also investigated 
for specific tasks~\cite{chenxing2021asynchronous_split,ayad2021autoencoder_split}.
%

In our work, we focus on reducing the communication for the classification problems with \textit{a large number of classes}, and propose a modification of top-$k$ sparsification, namely, \textit{randomized} top-$k$ sparsification (RandTopk).
To perform randomized top-$k$ sparsification on a $d$-dimensional vector, we first determine its top-$k$ elements.
Then we randomly select $k$ distinct elements such that: with probability $1 - \alpha$, the element is chosen from top-$k$ elements with equal chance, and with probability $\alpha$, the element is chosen from non-top-$k$ elements with equal chance.
Here, $\alpha\in[0, 1]$ is a hyperparameter to control the randomness.
The core idea behind randomized top-$k$ sparsification comes from the analysis of size reduction and top-$k$ sparsification.
We find that when the number of classes is small (e.g. $\le 10$), simply reducing the size of the cut layer can preserve model performance well.
However, this does not hold when the number of classes is large (e.g. $\ge 100$), since hidden features of different classes will be crowded within a low-dimensional manifold, making the model less smooth and generalize poorly.
Top-$k$ sparsification overcomes this problem by expanding the volume of the low-dimensional manifold.
However, it still faces local minimum problems, and usually fails to exploit the entire volume of the manifold due to unevenly selected neurons, i.e., some neurons are selected frequently while some are rarely selected.
RandTopk tackles both problems by adding randomness to top-$k$ sparsification.
First, the local minimum can be avoided by noisy gradients;
Second, since non-top-$k$ neurons also have chances to be selected during forward propagation, after training, the distribution of selected neurons is more balanced.

We conduct extensive experiments on different tasks, and compare RandTopk with top-$k$ sparsification, cut layer size reduction, quantization, and L1 regularization.
Empirical results show that RandTopk significantly outperforms top-$k$ sparsification and other compression methods when their compression ratios are on the same level.
We summarize our main contributions as follows:
\begin{itemize}[leftmargin=20pt]
    \item We investigate different compression methods for split learning to improve its communication efficiency, including cut layer size reduction, top-$k$ sparsification, quantization, and L1 regularization.
    \item Through the analysis of the cut layer size reduction and top-$k$ sparsification, we propose randomized top-$k$ sparsification, to overcome the convergence and generalization problems of top-$k$ sparsification.
    \item We conduct extensive experiments and show that randomized top-$k$ sparsification achieves the best model performance compared to other compression methods when their compression ratios are on the same level.
\end{itemize}

\section{Related Work}
\subsection{Reducing Communication for HFL}
A classic communication reduction method for HFL is the famous FedAvg scheme~\cite{mcmahan2017fedavg}, which reduces communication rounds for HFL by letting client train more rounds locally, and has now become the standard HFL scheme.
Based on this, studies aiming to reduce communication in HFL can be mainly divided into two classes, i.e., making convergence faster and compressing the model updates.
Methods of the former class usually accelerate convergence by making model updates of different clients more consistent~\cite{wanghongyi2020fedma,wangjianyu2020fednova,Karimireddy2020scaffold,jin2022accelerated}.
The latter class includes quantization, sparsification, and other compression methods.
The basic top-$k$ sparsification is well-studied both empirically and theoretically and is proven to be very effective~\cite{aji2017gradient_drop,stich2018sparse_memory}.
Combining sparsification and quantization can further reduce communication
~\cite{wenwei2017terngrad,sattler2019sparse_binary,sattler2019sparse_ternary}.
Other methods, such as using SVD to compress the model updates~\cite{wanghongyi2018atomo,vogels2019powerSGD} and using autoencoder as compressor~\cite{chandar2021autoencoder_communication}, are also investigated.

\subsection{Reducing Communication for VFL}
As for VFL, the studies on communication reduction are relatively few.
Castiglia et al. \shortcite{timothy2022compressed-vfl} studied the convergence using top-$k$ sparsification and quantization.
Other techniques like asynchronous training and autoencoder compressor are also used in specific tasks~\cite{chenxing2021asynchronous_split,ayad2021autoencoder_split}.
Chen et al.~\shortcite{chenxing2021asynchronous_split} use asynchronous training and quantization to reduce the communication for split learning.
Using asynchronous training, the bottom model is not updated every iteration, hence the top model no longer needs to send the gradient to the bottom model in every iteration.
The authors also show that this method neither enhances nor harms the privacy of split learning.
Ayad et al.~\shortcite{ayad2021autoencoder_split} use an autoencoder to compress the intermediate results.
However, this method requires injecting an autoencoder which has to be specifically designed for different tasks.

\section{Basic Compression for Split Learning}
\begin{table}[ht]
\footnotesize
    \centering
    \begin{tabular}{cl}
    \toprule
        Notation   & Definition    \\
    \midrule
        $M_b, M_t$                          & The bottom model and the top model \\
        $X$                                 & Input instance \\
        $O$                                 & Output of the bottom model \\
        $\hat Y, Y$                         & Model prediction and label \\
        $L(\hat Y, Y)$                      & Loss function \\
        $G_o$ or $G$                        & Gradient of the loss ($\partial L/\partial O_b$) \\
        $\mathsf{Comp}, \mathsf{Decomp}$    & Compression/Decompression function \\
        $\mathsf{Encode}, \mathsf{Decode}$  & Encode/Decode between values and bytes\\
        $d$                                 & Dimension of the bottom model output \\
        $k$                                 & \#Non-zero elements after sparsification\\
        $n$                                 & Number of classes\\
    \bottomrule
    \end{tabular}
    \caption{Notations \& Definitions}
    \label{table:notations}
\end{table}
In split learning, intermediate results are sent during each forward pass and backward pass.
Hence, the compression is performed on the intermediate results.
Specifically, in this paper, we only consider the compression on the \textit{instance level}, i.e., if there is a batch of instances, the compression method is applied to the instances' bottom model outputs inside the batch individually.
Let $\mathsf{Comp}(\cdot): \mathbb R^d \to \mathbb B^*$ and $\mathsf{Decomp}(\cdot): \mathbb B^*\to \mathbb R^d$ be the compression and corresponding decompression operator, where $\mathbb B^*$ is the bytes array of arbitrary length.
For simplicity, we denote $C[\cdot] \equiv \mathsf{Decomp}[\mathsf{Comp}(\cdot)]$. 
Note that the compression methods discussed in this paper usually are not lossless, hence $C[X]$ and $X$ are usually not equal.
Let $X$ be an input instance, and we assume that the output of $M_b$ is flattened to a $d$-dimensional vector, i.e., $\forall X, O_b = M_b(X) \in \mathbb R^d$.
Then we can define the compression for split learning during the forward pass as follows:
The feature owner sends $\mathsf{Comp}(O_b)$ to the label owner.
The label owner recovers $C[O_b] = \mathsf{Decomp}[\mathsf{Comp}(O_b)]$.
Notice that the backward pass or the loss function should also be modified according to different forward compression methods.
The notations we use are described in \Cref{table:notations}.

\subsection{Basic Compression Methods}
We now describe some basic compression methods for split learning.

\textbf{(Cut layer) size reduction}
is to reduce the size of the bottom model's output.
For example, consider a fully-connected network whose architecture is 1,000 (input)-100 (first hidden layer)-100 (second hidden layer)-10 (output).
Assuming that we split the model by its second hidden layer, the cut layer size is 100.
If we `slim' the network by changing its architecture to 1,000-100-10-10, then we achieve a 90\% compression ratio.
In case the network architecture is complicated and hard to modify, we can just apply a mask $M\in\mathbb R^d = (\underbrace{1,...,1}_\text{$k$ $1$'s},\underbrace{0,...,0}_\text{$d - k$ $0$'s})^T$ to $O_b$.
Hence, we can define the compression and decompression functions as
\begin{equation}
\small
\begin{split}
    \mathsf{Comp}(O) &= \mathsf{Encode}(o_1, ..., o_k),\\
    \mathsf{Decomp}(O') &= (\mathsf{Decode}(O'), \underbrace{0, ..., 0}_\text{$d - k$ $0$'s}).
\end{split}
\end{equation}
During the backward pass, we also apply the same compression and decompression function.
The reason is that the last $d - k$ entries of the bottom output are masked and the gradient w.r.t. them is meaningless to the bottom model.

\textbf{Quantization}
is to convert float values (usually 32-bit) to low-bit representations, which are widely used to improve the efficiency of neural networks~\cite{jacob2018quantization,amir2021quantization_survey}.
While there are many quantization methods, in this paper, we consider trivial uniform quantization.
Suppose the bottom output's range is $[o_\text{min}, o_\text{max}]$, and we want to quantize it into $b$-bit.
We generate $2^b$ bins, and the $n$-th bin has the range $[o_\text{min} + (o_\text{max} - o_\text{min})/2^b \cdot (n-1), o_\text{min} + (o_\text{max} - o_\text{min})/2^b \cdot n]$.
Hence, the compression and decompression functions are defined as 
\begin{equation}
\small
\begin{split}
    \mathsf{Comp}(O) &= \mathsf{Encode}\left(..., \left\lfloor\dfrac{o_i - o_\text{min}}{(o_\text{max} - o_\text{min})/2^b}\right\rfloor, ...\right), \\
    \mathsf{Decomp}(O') &= \left(..., o_\text{min} + (c_i + \dfrac12) \cdot [\dfrac{o_\text{max} - o_\text{min}}{2^b}], ...\right), \\
\end{split}
\end{equation}
where  $(c_1, ..., c_d) = \mathsf{Decode}(O')$.
Quantization can be applied on both the forward pass and the backward pass.
However, since the quantization of backward gradients significantly hurts the model performance and we mostly focus on the inference efficiency, hence, in this paper we apply quantization to the forward pass only.

\textbf{Top-$k$ sparsification} 
is to preserve top-$k$ elements in the vector, in terms of magnitude, while setting all other elements to zero.
We can define the compression and decompression functions as
\begin{equation}
\small
\begin{split}
    \mathsf{Comp}(O) &= \mathsf{Encode}(o_{j_1}, ..., o_{j_k},j_1,...,j_k), \\
    \mathsf{Decomp}(O') &= (..., I_{i\in J}\cdot o_i, ...), \\
\end{split}
\end{equation}
where $J = (j_1, ..., j_k)$ are the indices of largest-$k$ elements in $O$.
During the backward pass, the feature owner only needs the gradients on non-zero entries to update the bottom model.
Hence, we also apply the compression to the backward pass.
Moreover, during the backward pass, the top-$k$ indices are already maintained by the feature owner and hence need not be transferred.

\textbf{L1 regularization}
is widely used in different fields of machine learning to induce sparsity~\cite{tibshirani1996lasso,wright2008sparse_face,yin2012kernel}.
To make the bottom output sparse, we add the L1-norm of the bottom output to the loss, i.e.,
$L' = L + \lambda \sum_{i=1}^d |o_i|$,
where $L$ is the original loss, and $\lambda$ is a coefficient to control the sparsity.
Larger $\lambda$ tends to induce higher sparsity, but may also hurt the model performance more.
The compression and decompression function when using L1 regularization is the same as the top-$k$ sparsification case, the only difference is that the retrieved indices $J$ 
are non-zero indices rather than top-$k$ indices.
Note that in the backward propagation, no sparsification shall be applied.
\subsection{Compressed Size}
Different compression methods have different compressed sizes.
For simplicity, here we use (relative) compressed size to denote the ratio between compressed data and original data, which is the inverse of compression ratio.
For size reduction, the compressed size is simply $k/d$. Similarly, quantization has the compressed size $2^b/N$, where $N$ is the original values' bit-length, which is usually 32.
As for the top-$k$ sparsification and L1 regularization, although only $k$ values are preserved, we also have to send their indices during the forward pass.
Suppose that a single index needs $r$ bits to encode, then the compressed size of top-$k$ sparsification or L1 regularization is $k/d\cdot (1 + r/N)$.
In this paper, we consider the offset encoding that $r = \lceil \log_2 d \rceil$.
We conclude the compressed size of different methods in \Cref{tab:compression_ratio}.
\begin{table}[h]
\small
    \centering
    \begin{tabular}{ccc}
    \toprule
    \multirow{2}{*}{Method} & \multicolumn{2}{c}{Compressed size} \\ 
    \cmidrule(lr){2-3}
         & Forward & Backward \\ \midrule
    Cut layer size reduction    & $k/d$       & $k/d$ \\
    Quantization to $b$-bits    & $2^b/N$    & 1 \\
    Top-$k$ sparsification      & $k/d\cdot (1 + \lceil \log_2 d \rceil/N)$ & $k/d$ \\
    L1 regularization           & $k/d\cdot (1 + \lceil \log_2 d \rceil/N)$ & 1 \\
    \bottomrule
    \end{tabular}
    \caption{Compressed size of different methods, 
    where $k$ is the number of non-zero elements, 
    $d$ is the size of original bottom model output, 
    $b$ is the bit-length of quantized value,
    and $N$ is the original values' bit-length, which is usually 32.
    }
    \label{tab:compression_ratio}
\end{table}
\subsection{Summary}
From the above discussion, we can see that for size reduction and top-$k$ sparsification, we can explicitly set the compression ratio by deciding the portion of preserved 
elements.
Quantization methods can only achieve a maximum compression ratio of 32 if the original value is 32-bit.
As for L1 regularization, the compression ratio is hard to control.
It is hard to estimate the compression ratio until the training is finished, and the compression ratio even varies on different inputs.
By our experiments in \Cref{sec:exp}, we can see that quantization and L1 regularization fail to converge or are infeasible under many compression levels.
%





\section{Randomized Top-$k$ Sparsification for SL}
In this section, we propose our method: randomized top-$k$ sparsification (RandTopk).
The idea of RandTopk comes from the analysis of size reduction and top-$k$ sparsification, as they are all implemented by dropping some elements in the bottom model output.
In this section, we first demonstrate that top-$k$ is (theoretically) better than size reduction since it provides a larger feature space under the same compression ratio, which leads to better generalization.
Based on this, we introduce RandTopk and demonstrate its advantage over top-$k$ from two aspects: convergence and generalization.
\subsection{Analysis of Top-$k$ and Size Reduction}
\label{sec:topk}
We first demonstrate that if the cut layer's size is small and there are a large number of classes, the model will suffer from a huge generalization error because it becomes less smooth.
Then we illustrate that using top-$k$ sparsification, we can make the model much smoother than size reduction.

\paragraph{Larger margin, better generalization.}
Studies have shown that the generalization ability of neural networks is closely related to its smoothness~\cite{neysharbur2015norm_capacity,neysharbur2017generalization,gouk2021lipschitz_reg}.
Generally speaking, more smoothness leads to better generalization.
Here we consider the case that the cut layer is the last hidden layer, while the top model is a linear layer with softmax activation.
The model prediction is 
\begin{equation}
    \mathbf y = \text{Softmax}(\mathbf o\cdot \mathbf w_1, ..., \mathbf o\cdot \mathbf w_n),
\end{equation}
where $\mathbf o$ is the bottom model output, $\mathbf w_i$ is the weight (embedding, hidden feature) in last layer corresponding to $i$-th output neuron, and $n$ is the total number of classes (ignoring the bias).
When the model is trained well enough, if the input belongs to the $i$-th class, then $\mathbf o$ shall be close to $
\mathbf w_i$ and far from $\mathbf w_{j\ne i}$.
Let the \textit{minimum margin} (between two different classes' weights) $d_W$ be
$
    d_W = \min_{i \ne j} \Vert \mathbf w_i - \mathbf w_j \Vert_2,
$
the smoothness of the top model can be approximated as $\Vert \nabla_{\theta_t} \Vert_2 \approx c / d_W$, where $c$ is some constant.

Hence, we can use $d_W$ as an indicator for generalization ability.
Larger $d_W$ means embeddings of different classes are away from each other, leading to lower generalization error, and vice versa.

\paragraph{Top-$k$ has a larger margin due to a larger feature space.}
The minimum margin between different classes' weights is proportional to the volume of the \textit{feature space}, i.e., the space that $\mathbf w_i$ lies in.
Obviously, if we do not put any restriction on $\mathbf w_i$, the feature space always has an infinite measure.
To avoid this, notice that the Softmax result will not change if the weight/input is multiplied by a constant.
In other words, what matters is the direction of $\mathbf o$ and $\mathbf w$, instead of the magnitude.
Then we can assume $\Vert \mathbf  w_i \Vert^2 = 1$.
Hence, each $\mathbf w_i$ can have a ball $B_i$ of radius $d_W/2$, hence there are totally $n$ balls corresponding to $n$ classes.
Those balls are disjoint, thus the sum of those balls' volume is less than the volume of the entire feature space.
While those balls are $k$-dimensional, we further consider their intersection with the hypersphere $\Vert \mathbf w \Vert_2 = 1$.
This yields a $(k-1)$-dimensional manifold.
While we consider any $B_i$ is small, the manifold is close to a $(k-1)$-ball, whose volume is $\pi^{(k-1)/2}/\Gamma((k + 1) / 2)(d_W/2)^{k-1}$.

For the cut layer of size $k$, $\mathbf w_i$'s are $k$-dimensional.
Since the hypersphere $\Vert \mathbf w\Vert_2 = 1$ has a volume of $2\pi^{k/2}/\Gamma(k/2)$, we have 
\begin{equation}
    n\cdot\pi^{(k-1)/2}/\Gamma((k+1)/2) (d_W/2)^{k-1} \lesssim 2\pi^{k/2}/\Gamma(k/2).
\end{equation}
Approximation is used here because the volume is computed on a hypersphere rather than a flat Euclidean space.
Then we can estimate that, if all $\mathbf w_i$'s are almost uniformly distributed on the hypersphere, $d_W \approx 2\cdot(2/n\sqrt{k\pi/2})^{1/(k-1)}$.

While size reduction only selects the first $k$ dimensions of the bottom model output, top-$k$ sparsification has $d\choose k$ different selections, making the minimum distance $d \choose k$ times larger.
However, due to the extra communication of top-$k$ sparsification for sending the indices, we can only have $k' = \alpha k < k$ non-zero elements,
where $\alpha$ is considered to be larger than $1/2$ since we do not need 32 bits to encode the indices (otherwise the cut layer size will be up to $2^{32}$).
Comparing the minimum distance of size reduction and top-$k$ sparsification, 
and assume that $n^2 \ge 2\alpha k \pi$ (which is obvious in our scenario), 
we have
\begin{equation}
\label{eq:min-dist}
\begin{split}
    & \dfrac{d_W^\text{(top-$k$)}}{d_W^\text{(size reduction)}} \gtrsim {d\choose \alpha k} \cdot \dfrac{\sqrt{2\alpha^3k\pi}}{n}.
\end{split}
\end{equation}
It is obvious that if $n \le \sqrt{k/2} {d \choose \alpha k}$, \Cref{eq:min-dist} is much larger than $1$.
This condition holds in most scenarios since ${d \choose \alpha k}$ is a high-order polynomial of $d$.
Refer to Appendix D for details.

In conclusion, under the same compression level, using top-$k$ sparsification makes the feature space larger, and results in a smoother top model, finally leading to a smaller generalization error.

\subsection{Adding Randomness to Top-$k$}
\label{sec:RandTopk}
We show that by adding randomness to top-$k$ sparsification, the model converges better, and even generalizes better than top-$k$.
We add randomness to top-$k$ sparsification by selecting the neurons as follows:
\begin{equation}
\begin{split}
\small
 P(\text{select $i$-th neuron})
   =\begin{cases}
        (1 - \alpha) / N_1  & \text{if $i$-th neuron is top-$k$,}\\
        \alpha / N_2     & \text{otherwise.}
    \end{cases}
\end{split}
\end{equation}
The selection is performed $k$ times without replacement, and $N_1, N_2$ denote the numbers of remaining top-$k$/non-top-$k$ neurons.
We can see that $\alpha$ is a hyperparameter to control the degree of randomness. 
Larger $\alpha$ makes non-top-$k$ neurons more likely to be selected.
With $\alpha=0$, our method behaves exactly the same as top-$k$, and with $\alpha=1$, it becomes the Dropout.
Empirical analysis on $\alpha$ can be found in Appendix C.
Notice that randomness is only added during the training procedure.
For model inference, our method adopts the same behavior as top-$k$.
We will mention the proposed randomized top-$k$ sparsification as RandTopk for short.
We further argue the advantages of RandTopk as follows.

\begin{figure}[t]
    %
    \centering
    \includegraphics[width=0.9\linewidth]{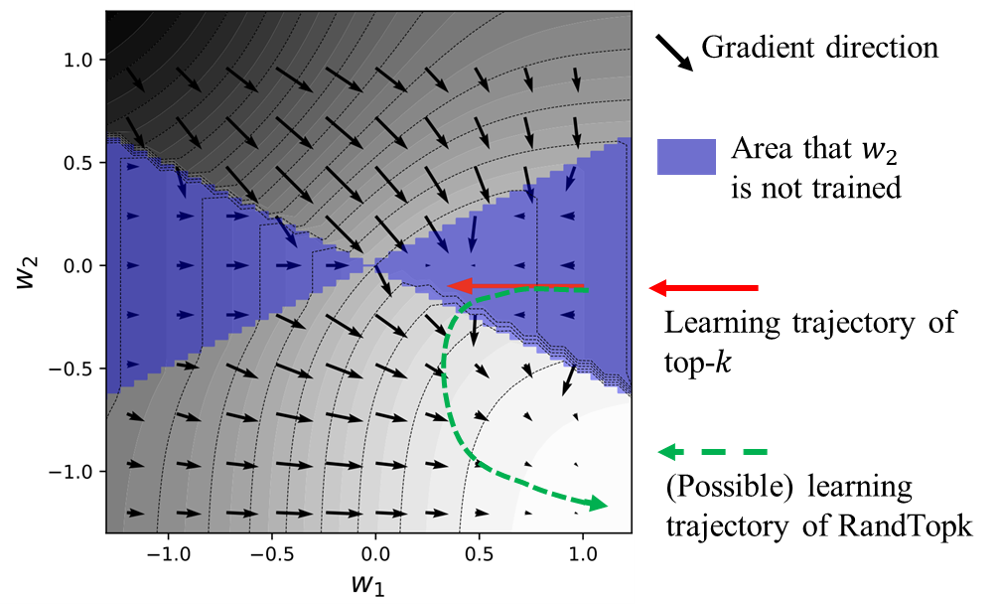}
    \caption{The loss surface, gradient field, and learning trajectory of the toy example.}
    \label{fig:topk-example-gradient-field}
\end{figure}

\paragraph{Better convergence.}
One concern about top-$k$ sparsification is that the model may be stuck at local minimums since those small neurons are not trained.
RandTopk overcomes this problem naturally since its randomness, which makes small neurons get opportunities to be trained.
Hence, the model will converge faster and to a better minimum.

To better illustrate this, we give the following toy example:
Assume that we are learning the concept 
\begin{equation}
    f: (x_1, x_2) \rightarrow \text{Sign}(x_1 - x_2),
\end{equation}
with a simple split logistic model 
\begin{equation}
\small
\begin{split}
       & M_h: (x_1, x_2)\rightarrow (o_1, o_2) = (w_1x_1, w_2x_2), \\
       & M_t: (o_1, o_2)\rightarrow \text{Tanh}(o_1 + o_2).
\end{split}
\end{equation}
The initial weights of the model are $w_1 = 1, w_2 = -0.1$,
and two samples are provided: $\mathbf x_1 = (1, 0), y_1 = 1$ and $\mathbf x_2 = (0.5, 1), y_2 = -1$.
It is obvious that if using top-$k$, $w_1 = +\infty, w_2 = -\infty$ are the optimal weights.
However, the optimal weights can never be achieved since $o_2$ is always masked by top-$k$ sparsification, and $w_2$ is never trained.
To make it clearer, we plot the loss surface and the gradient field of this example in \Cref{fig:topk-example-gradient-field}, and we fill the area where $w_2$ cannot be trained with blue.
We can see that because of the non-continuity of gradient introduced by top-$k$, there is a bad local minimum in the blue area, and our training (red arrow) ends up there.
Adding randomness to top-$k$, i.e., give non-top-$k$ neurons a chance to be selected, the problem can be solved instantly since $w_2$ will also be trained.

\paragraph{Better generalization.}
As discussed above in \Cref{sec:topk}, top-$k$ sparsification generalizes better since it has a larger feature space (under the same compression ratio) brought by total $d \choose k$ hyperspheres.
However, training with top-$k$ sparsification cannot fully exploit the larger feature space, since the neurons are unevenly selected.
Some neurons are constantly being selected with different input samples, while some are rarely selected.
Those rarely selected neurons usually remain small for most input samples and are also rarely selected and trained throughout the training process.
Consequently, the model cannot fully exploit the $d \choose k$ hyperspheres brought by top-$k$.
For example, if there are $d'$ neurons that never become top-$k$, then the feature space is reduced to ${{d - d'} \choose k} < {d \choose k}$ hyperspheres.
The margins between different classes are also smaller.
RandTopk overcomes this problem since the small neurons are also selected.
If some small neurons have the potential to become top-$k$ (w.r.t. certain input examples), RandTopk can provide them a chance to be large.
Thus, the model can explore a larger space in those hyperspheres, and the margin will be larger, finally leading to a better generalization.

\begin{table*}[h]
\footnotesize
\centering
\setlength\tabcolsep{4px}
\begin{tabular}{@{}ccccccc@{}}
\toprule
Task                                                                                        & Comp. & RandTopk                           & Topk                      & Size reduction             & Quantization                 & L1 regularization                              \\ \midrule
\multirow{3}{*}{\begin{tabular}[c]{@{}c@{}}CIFAR-100\\      67.20 (0.71)\end{tabular}}      & High        & \textbf{65.25 (0.54) / 2.86}       & {\ul 62.23 (0.55) / 2.86} & 55.52 (0.69) / 3.13       & -                            & -                                   \\ \cmidrule(l){2-7} 
                                                                                            & Medium      & \textbf{65.83 (0.56) / 5.71}       & {\ul 61.56 (1.26) / 5.71} & 60.43 (0.33) / 6.25       & 53.56 (0.52) / 6.25          & {\ul 62.11  (0.56) / 8.41 (0.49)}   \\ \cmidrule(l){2-7} 
                                                                                            & Low         & {\ul 65.98 (0.21) / 12.38}         & 62.11 (0.71) / 12.38      & 62.93 (0.57) / 12.5       & \textbf{66.01 (0.23) / 12.5} & 63.87 (0.36) / 19.5 (2.25)          \\ \midrule
\multirow{3}{*}{\begin{tabular}[c]{@{}c@{}}YooChoose\\      63.57 (0.57)\end{tabular}}      & High        & \textbf{60.29 (0.05) / 0.85}       & {\ul 60.28 (0.41) / 0.85} & 50.71 (1.51) / 1.00       & -                            & -                                   \\ \cmidrule(l){2-7} 
                                                                                            & Medium      & \textbf{64.55 (0.16) / 1.71}       & {\ul 63.81 (0.36) / 1.71} & 62.20 (0.50) / 2.00       & -                            & -                                   \\ \cmidrule(l){2-7} 
                                                                                            & Low         & \textbf{66.88 (0.13) / 3.84}       & {\ul 66.12 (0.44) / 3.84} & {\ul 66.12 (0.09) / 4.00} & 64.69 (0.21) / 3.13          & 61.48 (5.28) / 3.01 (1.12)          \\ \midrule
\multirow{4}{*}{\begin{tabular}[c]{@{}c@{}}DBPedia\\      93.11 (0.13)\end{tabular}}        & High+       & \textbf{84.88 (0.47) / 0.44}       & {\ul 83.04 (0.45) / 0.44} & 64.80 (1.09) / 0.50       & -                            & -                                   \\ \cmidrule(l){2-7} 
                                                                                            & High        & \textbf{88.01 (0.23) / 0.88}       & {\ul 85.49 (0.35) / 0.88} & 78.57 (0.53) / 1.00       & -                            & 81.35 (0.68) / 1.08 (0.02)          \\ \cmidrule(l){2-7} 
                                                                                            & Medium      & \textbf{90.50 (0.11) / 1.97}       & 87.74 (0.35) / 1.97       & 86.42 (0.26) / 2.00       & -                            & {\ul 87.88 (0.17) / 0.93 (0.01)}    \\ \cmidrule(l){2-7} 
                                                                                            & Low         & {\ul 91.59 (0.50) / 3.06}          & 90.05 (0.14) / 3.06       & 88.38 (0.10) / 3.00       & 91.20 (0.20) / 6.25          & \textbf{93.11 (0.05) / 5.31 (0.24)} \\ \midrule
\multirow{3}{*}{\begin{tabular}[c]{@{}c@{}}TinyImagenet\\      53.11 (0.18)\end{tabular}}   & High        & \textbf{50.83 (0.81) / 0.21}       & 48.36 (0.09) / 0.21       & 35.46 (0.98) / 0.23       & -                            & -                                   \\ \cmidrule(l){2-7} 
                                                                                            & Medium      & \textbf{51.75 (0.04) / 0.42}       & {\ul 47.24 (0.09) / 0.42} & 45.66 (0.29) / 0.47       & -                            &  -                                 \\ \cmidrule(l){2-7} 
                                                                                            & Low         & \textbf{51.16 (0.14) / 0.94}       & 45.50 (0.19) / 0.94       & {\ul 48.87 (0.14) / 0.14} & -                            & -                                   \\ \midrule
\multirow{3}{*}{\begin{tabular}[c]{@{}c@{}}TinyImagenet-p\\      75.18 (0.29)\end{tabular}} & High        & \textbf{71.09 (0.14) / 0.21}       & {\ul 70.86 (0.07) / 0.21} & 59.23 (0.83) / 0.23       & -                            & -                                   \\ \cmidrule(l){2-7} 
                                                                                            & Medium      & \textbf{72.15 (0.15) / 0.42}       & {\ul 71.19 (0.54) / 0.42} & 66.52 (0.25) / 0.47       & -                            & -                                   \\ \cmidrule(l){2-7} 
                                                                                            & Low         & \textbf{73.83 (0.20) / 0.94}       & {\ul 72.52 (0.69) / 0.94} & 68.67 (0.16) / 0.94       & -                           & 67.82 (1.15) / 1.24 (0.04)          \\ \bottomrule
\end{tabular}
\caption{Accuracy and Compressed size, in ``Accuracy (standard deviation)/Compressed size (standard deviation, if not zero)'' format. 
    We assume the original communication size (non-compression case) to be 100.
    The accuracy of vanilla split training is under the dataset name.
    The best result is marked in bold, and the second-best result is underlined.
    The methods that fail to converge or are infeasible under the given compression level are omitted.
    }
\label{table:main-result}
\end{table*}

\subsection{Discussion on Privacy}
RandTopk has better input privacy than vanilla split learning since sparsification is performed on the bottom model output.
This is because a large portion of the elements in the bottom model output are zeroed out, it contains less information than the original non-sparsified output, which has also been demonstrated by previous studies~\cite{zhuligeng2020deep_leakage,fuchong2022label_inference}.
We also provide the results on input reconstruction result in Appendix B.
However, as split learning is vulnerable to label inference attacks~\cite{fuchong2022label_inference}, RandTopk cannot preserve more label privacy.
Hence, it is suitable for scenarios where the number of classes is large such that it is impossible for label inference attacks.
For example, a face recognition model with thousands of different faces on user devices, or a recommendation model running on the user's browser.


\section{Experiments}
\label{sec:exp}
\subsection{Settings}
\label{sec:exp-setting}
\begin{table}[h]
\small
    \centering
    \begin{tabular}{ccc} \toprule
        Dataset         & \#Classes& Dim. of last layer    \\ \midrule 
        CIFAR-100       &   100     & 128                   \\
        YooChoose       &   18,153  & 300                   \\
        DBPedia         &   219     & 600                   \\
        Tiny-Imagenet   &   200     & 1,280                  \\ \bottomrule
    \end{tabular}
        \caption{Dataset details.}
    \label{table:datasets}
\end{table}
We perform the experiments of different compression methods on four datasets, i.e., CIFAR-100~\cite{krizhevsky2009cifar}, YooChoose~\cite{ben2015yoochoose}, DBPedia~\cite{2007dbpedia}, and Tiny-Imagenet~\cite{tiny-imagenet}, using Resnet-20~\cite{hekaiming2016resnet}, GRU4Rec~\cite{hidasi2016gru4rec}, TextCNN~\cite{kimyoon2014textcnn}, and EfficientNet-b0~\cite{tanmingxing2019efficientnet}, respectively. 
For CIFAR-100 and Tiny-Imagenet, data augmentation including random cropping and flipping is used.
For YooChoose, we only use the latest 1/64 subset and apply the same data preprocessing as in \cite{hidasi2016gru4rec} while the loss function is changed to cross-entropy. The size of GRU layer is set to 300.
For TextCNN, we set kernel sizes to [3,4,5] and use the Glove word embeddings~
\cite{pennington2014glove} as initialization.
For Tiny-Imagenet, we train the model 
either from scratch (no suffix) or starting with pre-trained weights (marked by the suffix ``-p'').
The details of the datasets are provided in \Cref{table:datasets}.

We split all the models by their last layer, and apply different compression methods with different compression ratios.
All experiments are repeated 5 times, and the standard deviation is reported on the results.
We report the results on accuracy vs. compression at the inference phase, convergence speed during training, in terms of epochs and communication size, and further analysis on the RandTopk corresponding to our theoretic claims in \Cref{sec:RandTopk}.
Due to space limits, full experiment results, input reconstruction attacks, and analysis on the randomness coefficient $\alpha$ are placed in Appendix A, B, C, respectively.
\begin{figure*}[h]
    \centering
    \includegraphics[width=0.95\linewidth]{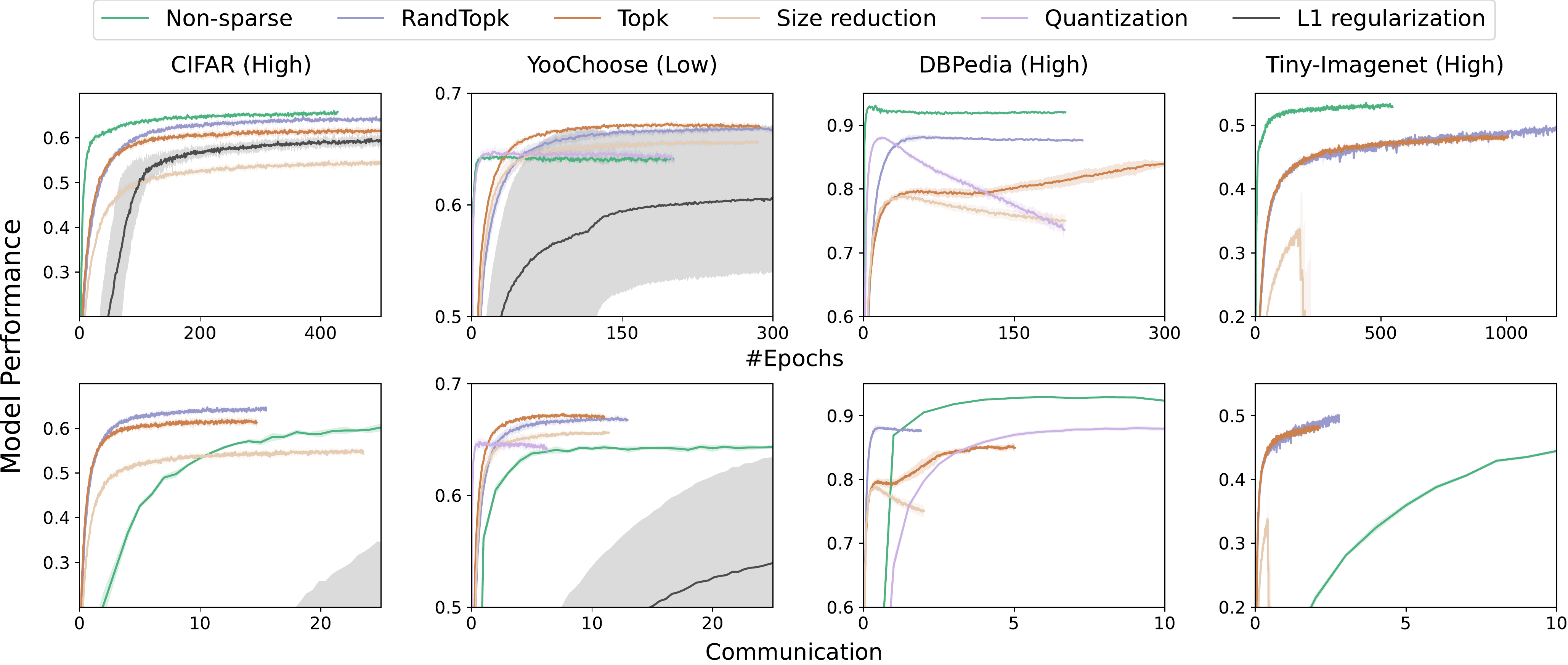}
    \caption{
    Convergence speed of different methods under certain compression levels.
    Compressed rates can be found in \Cref{table:main-result}.
    Top row: accuracy vs. \#epochs. Bottom row: Accuracy vs. communication 
    (total size of the message transferred, vanilla split learning in one epoch = 1). 
    Methods that are not applicable/fail to converge under the given compression level are omitted. }
    \label{fig:convergence}
\end{figure*}

\begin{figure}[h]
    \centering
    \begin{subfigure}{0.45\linewidth}
        \includegraphics[width=1\linewidth]{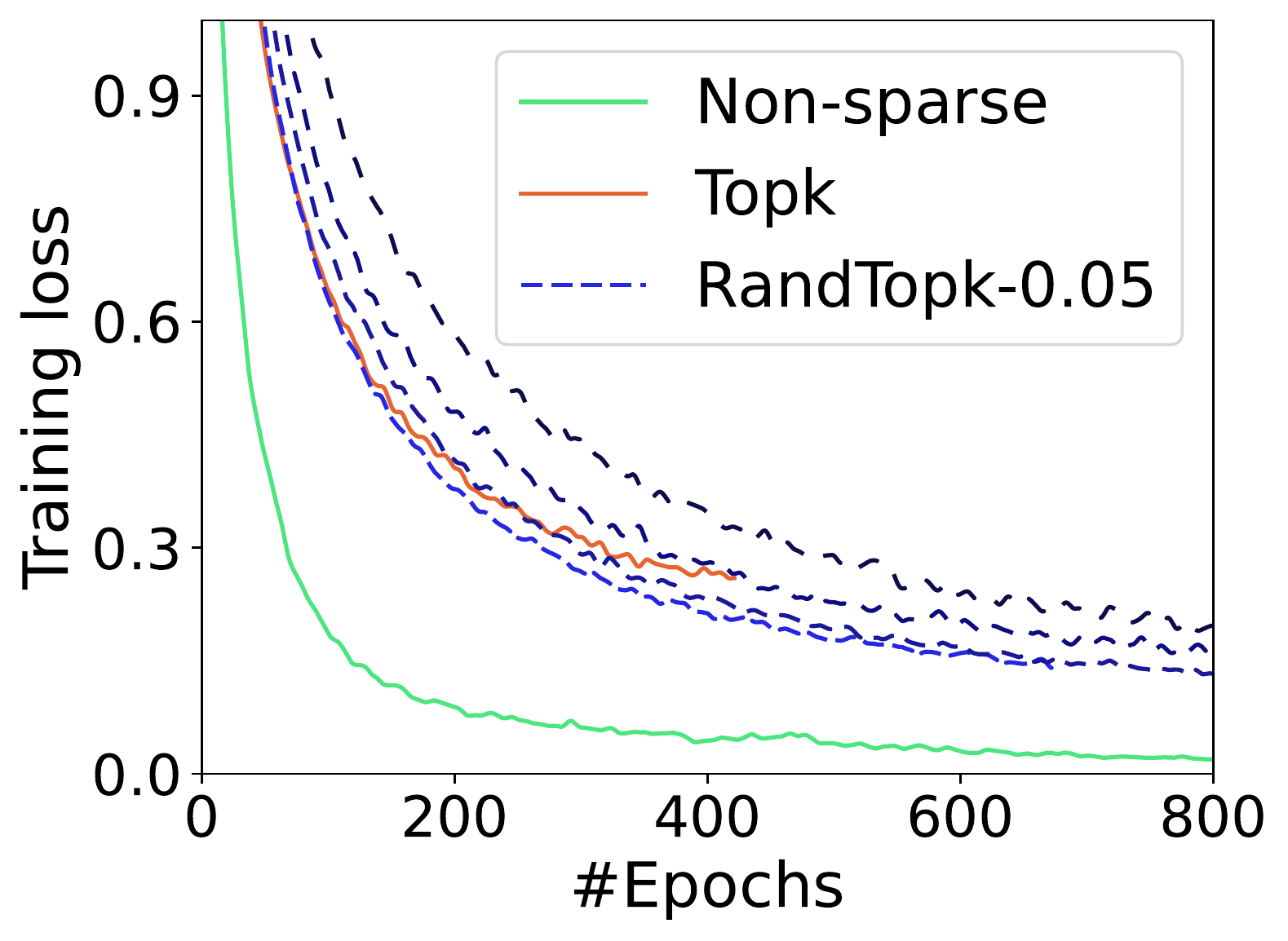}
        \subcaption{Loss on the training set.}
        \label{fig:cifar-trainloss}
    \end{subfigure}
    \begin{subfigure}{0.48\linewidth}
        \includegraphics[width=1\linewidth]{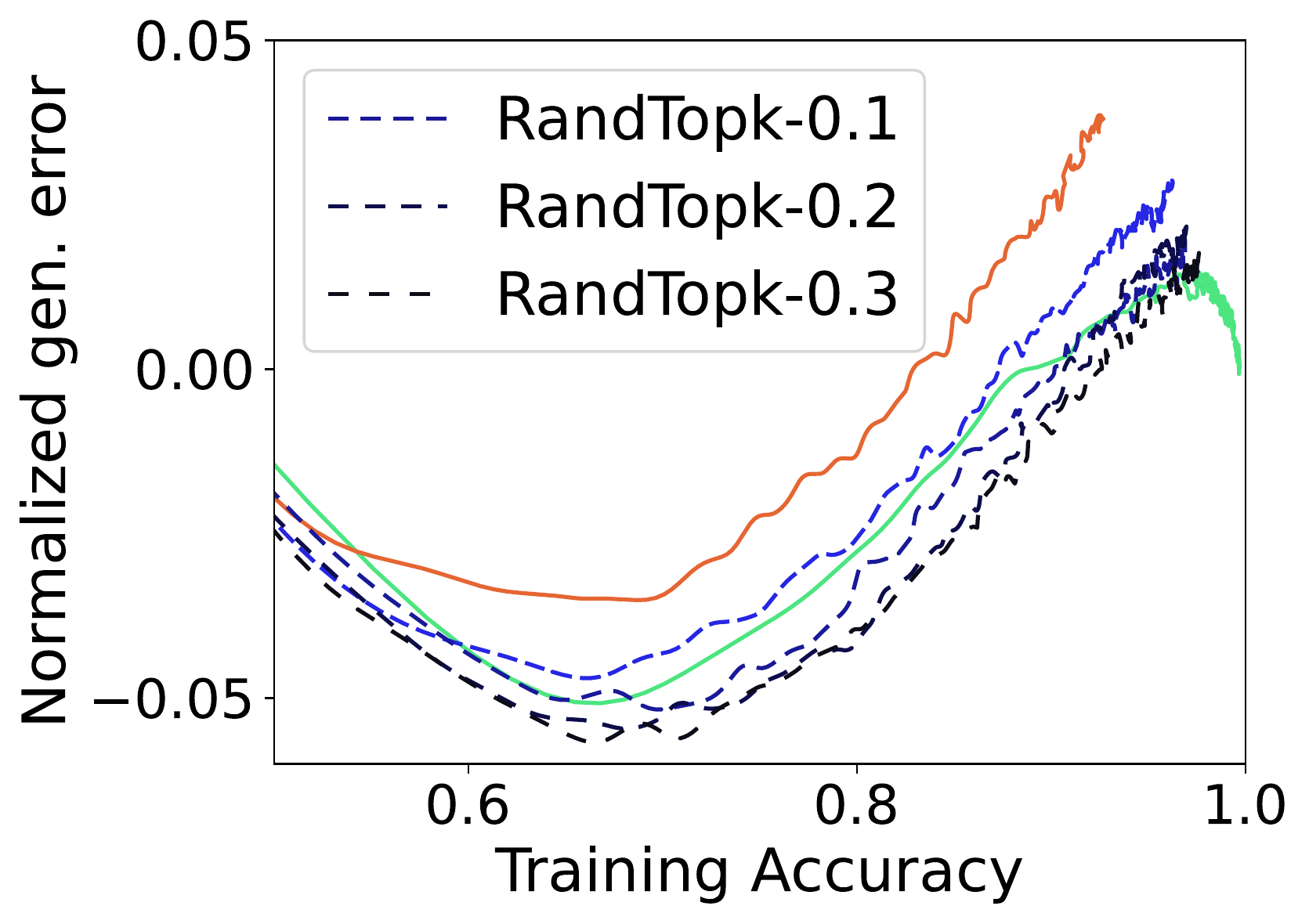}
        \subcaption{Generalization error.}
        \label{fig:cifar-generror}
    \end{subfigure}
    \caption{Training loss and generalization error on CIFAR-100.
            The compressed size is 2.86\%.}
    \label{fig:cifar-further}
\end{figure}

\subsection{Compressed Size vs. Performance}
We report the test accuracy (hit ratio@20 for YooChoose) and corresponding compressed size for inference on different tasks and compression levels in \Cref{table:main-result}.
Each experiment has been repeated 3 times for TinyImagenet and 5 times for other tasks for more precise results.
We set $\alpha$ to $0.1$ for all tasks except for YooChoose, where $\alpha = 0.05$.

We can see that RandTopk achieves the best performance in almost all tasks and all compression levels, and are usually significantly better than the second-best method.
Moreover, RandTopk's performance is often very near to the non-sparse case (vanilla SL)  even with a high compression ratio.
RandTopk surpasses the non-compression case in the YooChoose task, which we think is caused by its regularization effect. 

We also notice that quantization and L1 regularization are not applicable under many compression levels.
For quantization, it can only achieve a maximum compression ratio of 32, and 1-bit quantization usually fails to converge.
L1 regularization, on the other side, makes training difficult and usually fails to converge.

\subsection{Convergence Speed}
Although our main purpose is to reduce communication at the inference phase, we also report the convergence speed during training in \Cref{fig:convergence}.
The convergence speed is measured in two terms: the number of epochs and the communication size.

We can see that non-sparse (vanilla SL) training takes the least epochs to converge, while RandTopk and other compression methods converge slower, but the difference is not large to the magnitude level.
As for the overall communication to converge or reach a certain accuracy level, almost all compression methods outperforms the non-sparse training.
RandTopk performs best in terms of convergence speed and performance among all other methods.

\subsection{Further Analysis on RandTopk}
\label{app:further}
To verify our theoretical analysis in \Cref{sec:RandTopk}, we provide the results on convergence speed, generalization error, and the distribution of top-$k$ neurons during training using CIFAR-100, to demonstrate the advantage of RandTopk.

\paragraph{Faster convergence.}
We plot the loss curve of top-$k$ and RandTopk with different $\alpha$ in \Cref{fig:cifar-further}\subref{fig:cifar-trainloss}.
We can see that at first top-$k$ sparsification converges faster, then gradually slows.
With a small $\alpha$, RandTopk converges faster and reaches a smaller loss value.
When $\alpha$ becomes larger, the convergence at the beginning is slower, however, later it will exceed top-$k$.
The result verifies our argument that RandTopk can help avoid local minimums that top-$k$ may get stuck at.

\paragraph{Better generalization.}
We report the generalization error in \Cref{fig:cifar-further}\subref{fig:cifar-generror}.
We notice that for all methods, the generalization error can be approximated by $0.5\times\text{train acc.} - 0.2$.
Hence, the y-axis is set to be the difference between generalization error and this value for better illustration.
We can see that when the train set accuracy is the same, RandTopk has a significantly lower generalization error than top-$k$, and larger $\alpha$ further decreases the generalization error.
This supports our argument that RandTopk generalizes better.

\begin{figure}[t]
    \centering
    \includegraphics[width=1\linewidth]{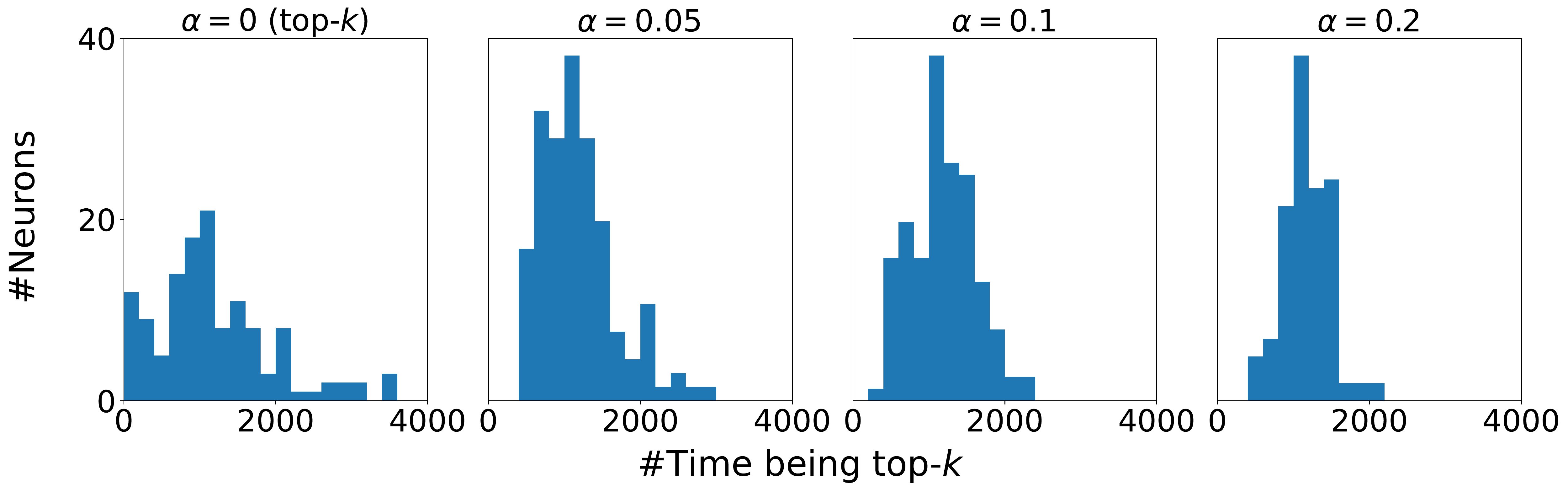}
    \caption{Distribution of top-$k$ neurons during inference.}
    \label{fig:topk-dist}
\end{figure}

\paragraph{Distribution of top-$k$ neurons.}
We report the histogram of the times a neuron become top-$k$.
More specifically, after model training is finished, we iterate through the train set and record the top-$k$ neurons for each input example.
Notice that we perform this experiment on the inference phase, hence RandTopk also behaves deterministically like top-$k$.
As we can see in \Cref{fig:topk-dist}, if trained with top-$k$ sparsification, some neurons will become top-$k$ neurons very often, while some rarely become top-$k$.
More specifically, some neurons become top-$k$ for less than 500 times, while some neurons become top-$k$ for more than 3,500 times. 
Using RandTopk, the distribution is more balanced, e.g., no neuron becomes top-$k$ more than 3,000 times or less than 250 times.
Larger $\alpha$ makes the distribution more balanced.
The results verify our argument that RandTopk can make use of the expanded feature space better.

\section{Conclusion}
\label{sec:conlusion}
In this paper, we investigate the use of different compression methods in split learning.
By the analysis of the size reduction method and top-$k$ sparsification, 
we further propose randomized top-$k$ sparsification, 
which strengthens top-$k$'s advantage on generalization ability while avoiding its disadvantage on convergence.
Experiments on multiple datasets and different kinds of models show that randomized top-$k$ is superior to other sparsification methods in terms of model performance and compression ratio.
%
%
However, there are still some issues worth studying in the future.
On the one hand, the label privacy remains to be a problem.
%
Moreover, further research on compression for split learning is needed, for example, combining quantization and sparsification can be promising.

\section*{Acknowledgements}
This work was supported in part by the “Pioneer” and “Leading Goose” R\&D Program of Zhejiang (No. 2022C01126), and Leading Expert of “Ten Thousands Talent Program” of Zhejiang Province (No.2021R52001).

{
\bibliographystyle{named}
\bibliography{ref}
}

\appendix
\setlength{\textfloatsep}{3pt}
\setlength{\floatsep}{0pt}
\setlength{\intextsep}{10pt}

\section{Detailed results on accuracy and compressed size}\
\label{app:detail}
We provide all results on accuracy and compressed size of all four tasks, in \Cref{table:cifar100}, \ref{table:yoochoose}, \ref{table:dbpedia} and \ref{table:tiny-imagenet}.
The coefficient of L1-regularization and RandTopk ($\alpha$) is marked.
For example, L1-0.001 means the regularization coefficient is 0.001, and RandTopk-0.1 means RandTopk with $\alpha = 0.1$.
\section{Input privacy}
\label{app:sec}
\begin{figure}[b]
    \centering
    \includegraphics[width=\linewidth]{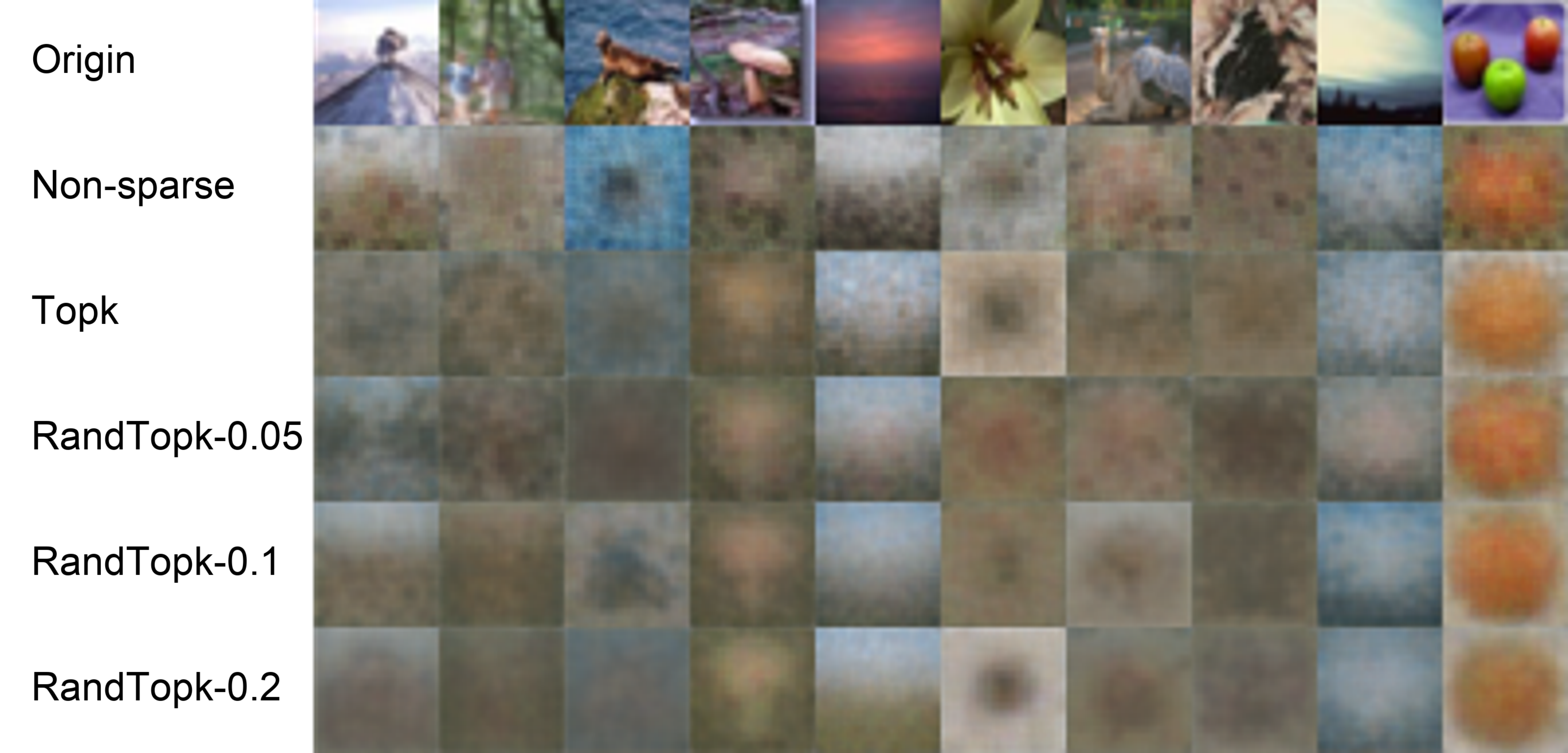}
    \caption{Inversion attack on CIFAR-100}
    \label{fig:attack}
\end{figure}

\begin{figure}[h!]
    \centering
    \includegraphics[width=1\linewidth]{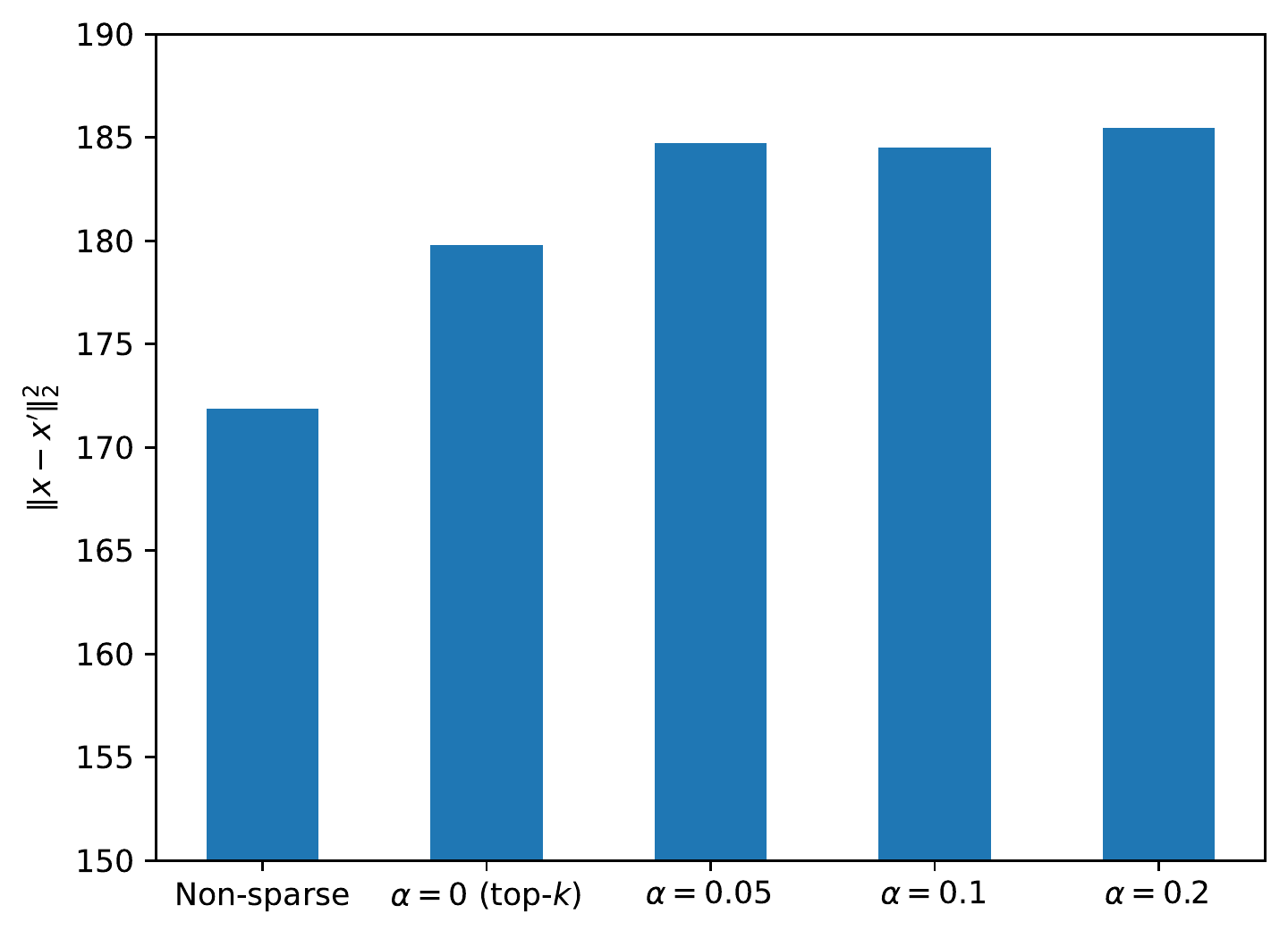}
    \caption{Reconstruction error of RandTopk on CIFAR-100}
    \label{fig:reconstruction-error}
\end{figure}

We also simulate the input inversion attack (i.e., inferring input from the bottom model output) on CIFAR-100, to show the privacy-preserving power of RandTopk.
We compared the non-sparse case (vanilla SL), with top-$k$ and RandTopk ($\alpha\in\{0.05, 0.1, 0.2\}$) sparsification of compressed size $2.86\%$, i.e., 3 out of 128 elements are preserved.
We train a generator network (made of Conv and DeConv layers) from the train set, whose input is the bottom model output and label is the original input image.
After training for 1,000 epochs, we use the test set to conduct the inversion attack.
The reconstruction error (squared loss) is reported in \Cref{fig:reconstruction-error}.
The original images and the generated images are shown in \Cref{fig:attack}.
We can see that using RandTopk, whatever $\alpha$ is, the reconstruction error is larger than top-$k$, and also much larger than the non-sparse case.
The reconstructed images are also very blurry and contain little information about the original input.

\section{Impact of $\alpha$}
\label{app:alpha}
As described in \Cref{sec:RandTopk}, $\alpha$ is used to control the randomness of RandTopk, and will potentially influence the model performance.
Hence, we report the results of using different $\alpha$ on the CIFAR-100 and YooChoose datasets.

We can see that, for the CIFAR-100 task, all choices of $\alpha\in\{0.05, 0.1, 0.2, 0.3\}$ perform significantly better than the top-$k$ (i.e., $\alpha = 0$), while $\alpha = 0.1$ achieves the best performance.
As for the YooChoose task, the best choice of $\alpha$ is around $0.05 \sim 0.1$.
Larger $\alpha$ leads to a significant performance drop, making it even worse than top-$k$.
Generally, too large $\alpha$ will introduce too much noise and damage the model performance, while $\alpha$ around $0.05\sim 0.1$ achieves good performance.

\begin{figure}[h!]
    \centering
    \begin{subfigure}{1\linewidth}
        \includegraphics[width=1\linewidth]{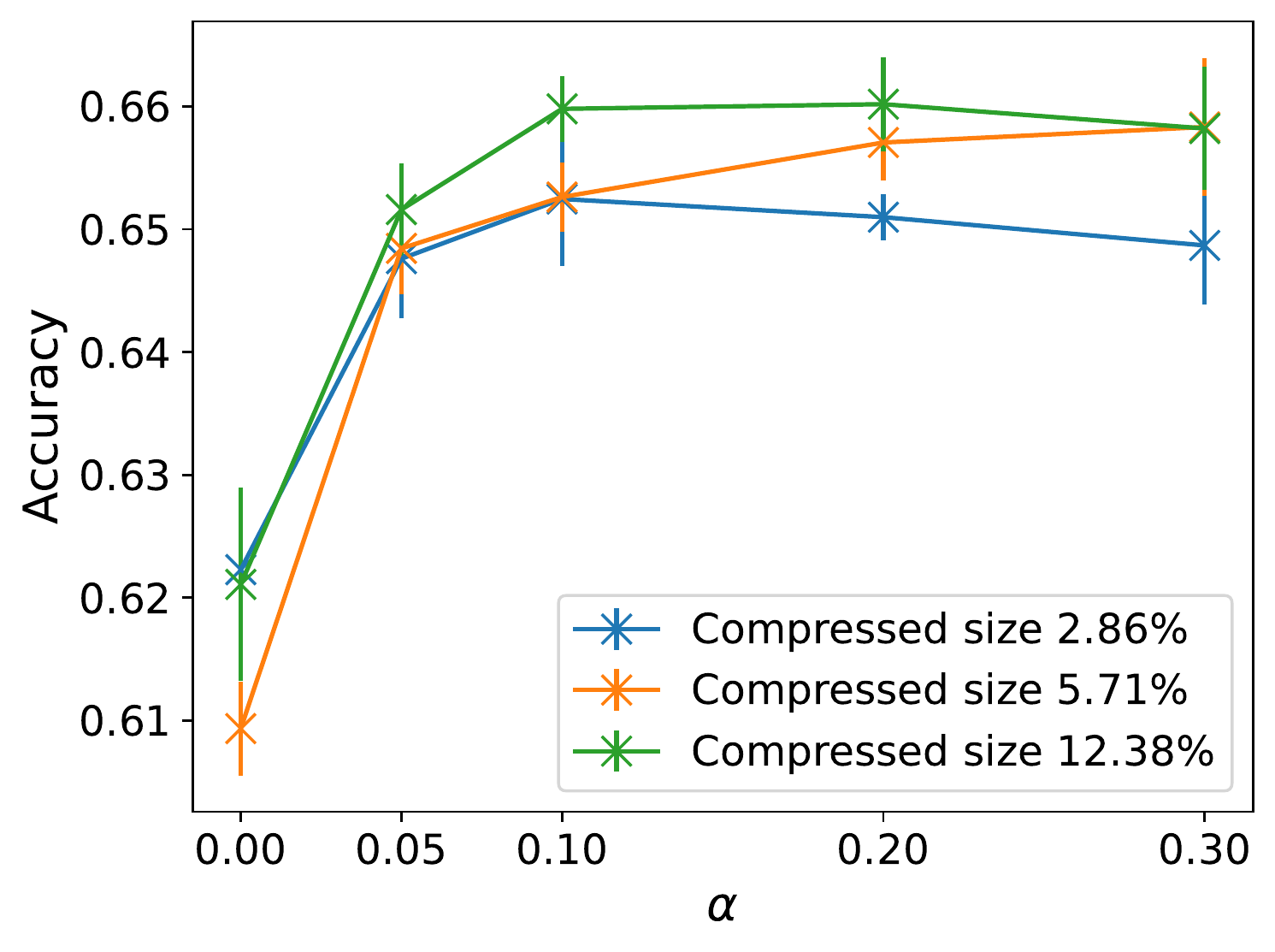}
        \caption{CIFAR.}
    \end{subfigure}
    \begin{subfigure}{1\linewidth}
        \includegraphics[width=1\linewidth]{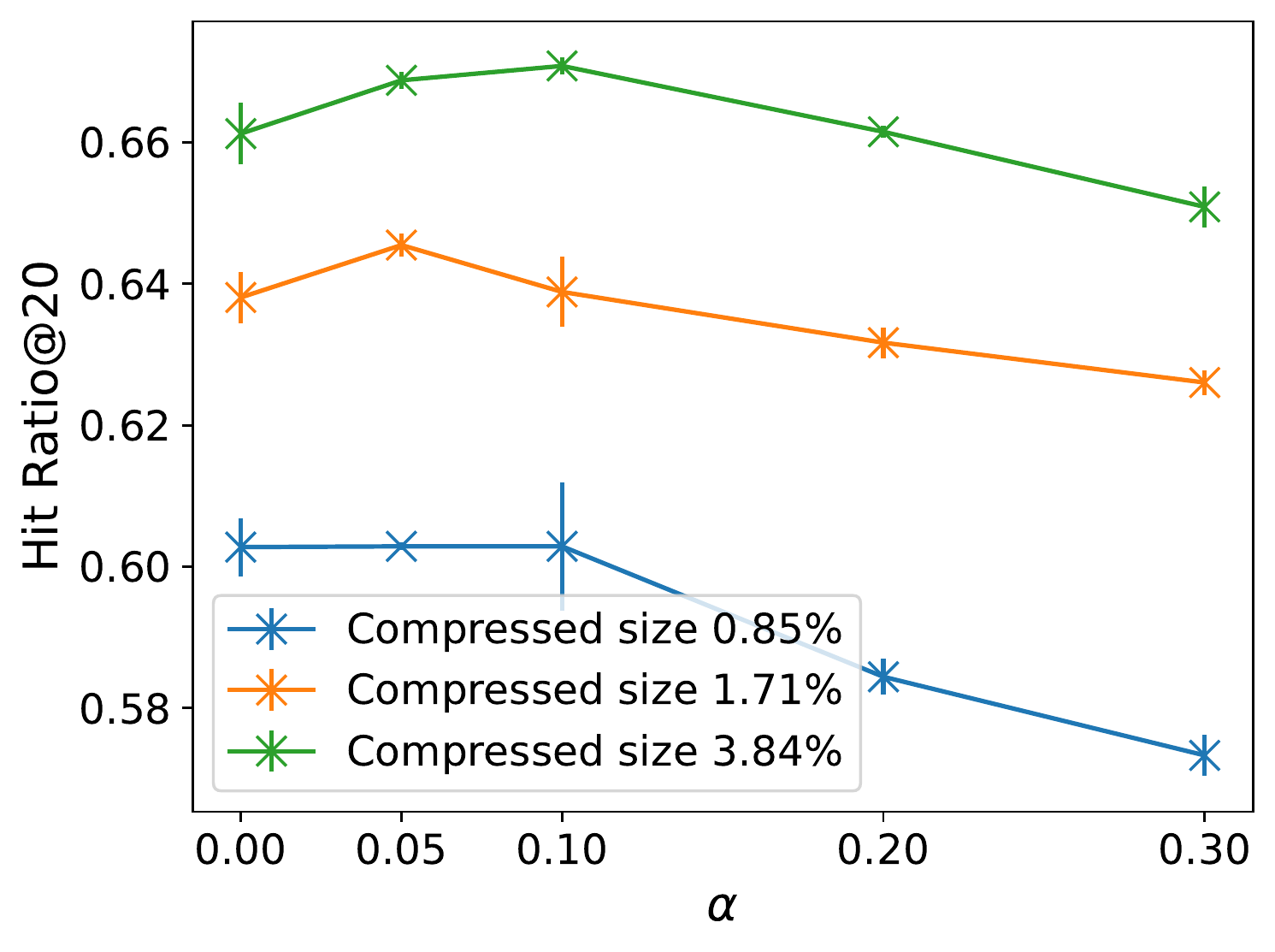}
        \caption{YooChoose.}
    \end{subfigure}
    \caption{Test accuracy on different $\alpha$.}
\end{figure}

\begin{table}[t]
\caption{Test accuracy and compression ratio on CIFAR-100.}
\label{table:cifar100}
\footnotesize
\centering
\begin{tabular}{@{}ccc@{}}
\toprule
Method            & Compressed size (\%) & Test accuracy        \\ \midrule
No compression    & 100                  & 67.20 (0.71)         \\ \midrule
Topk              & 2.86                 & 62.23 (0.55)         \\
RandTopk-0.05     & 2.86                 & 64.76 (0.49)         \\
RandTopk-0.1      & 2.86                 & 65.25 (0.54)         \\
RandTopk-0.2      & 2.86                 & 65.10 (0.19)         \\
RandTopk-0.3      & 2.86                 & 64.87 (0.48)         \\
Size reduction    & 3.13                 & 55.52 (0.69)         \\ 
\midrule
L1-0.001          & 5.45 (0.29)          & 60.96 (0.67)         \\
Topk              & 5.71                 & 61.56 (1.26)         \\
RandTopk-0.05     & 5.71                 & 64.84 (0.37)         \\
RandTopk-0.1      & 5.71                 & 65.26 (0.28)         \\
RandTopk-0.2      & 5.71                 & 65.71 (0.31)         \\
RandTopk-0.3      & 5.71                 & 65.83 (0.56)         \\
Size reduction    & 6.25                 & 60.43 (0.33)         \\
Quantization-2bit & 6.25                 & 53.56 (0.52)         \\
L1-0.0005         & 8.41 (0.49)          & 62.11 (0.56)         \\ 
\midrule
Topk              & 12.38                & 62.11 (0.79)         \\
RandTopk-0.05     & 12.38                & 65.16 (0.37)         \\
RandTopk-0.1      & 12.38                & 65.98 (0.27)         \\
RandTopk-0.2      & 12.38                & 66.02 (0.38)         \\
RandTopk-0.3      & 12.38                & 65.98 (0.27)         \\
Size reduction    & 12.5                 & 62.93 (0.57)         \\
Quantization-4bit & 12.5                 & 66.01 (0.23)         \\
L1-0.0002         & 19.5 (2.25)          & 63.87 (0.36)         \\ \bottomrule
\end{tabular}
\end{table}
\begin{table}[h!]
\caption{Test accuracy and compression ratio on YooChoose.}
\label{table:yoochoose}
\centering
\footnotesize
\begin{tabular}{@{}ccc@{}}
\toprule
Method            & Compressed size (\%) & Test hr@20            \\ \midrule
No compression    & 100                  & 63.57 (0.57)          \\ 
\midrule
Topk              & 0.85                 & 60.28 (0.41)          \\
RandTopk-0.05     & 0.85                 & 60.29 (0.05)          \\
RandTopk-0.1      & 0.85                 & 60.29 (0.91)          \\
RandTopk-0.2      & 0.85                 & 58.44 (0.25)          \\
RandTopk-0.3      & 0.85                 & 57.33 (0.29)          \\
Size reduction    & 1.00                 & 50.73 (1.51)          \\ 
\midrule
Topk              & 1.71                 & 63.81 (0.36)          \\
RandTopk-0.05     & 1.71                 & 64.55 (0.16)          \\
RandTopk-0.1      & 1.71                 & 63.89 (0.50)          \\
RandTopk-0.2      & 1.71                 & 63.17 (0.22)          \\
RandTopk-0.3      & 1.71                 & 62.60 (0.18)          \\
Size reduction    & 2.00                 & 62.20 (0.50)          \\ 
\midrule
L1-0.002          & 3.01 (1.12)          & 61.48 (5.28)          \\
Quantization-1bit & 3.13                 & 64.69 (0.21)          \\
Topk              & 3.84                 & 66.12 (0.44)          \\
RandTopk-0.05     & 3.84                 & 66.88 (0.13)          \\
RandTopk-0.1      & 3.84                 & 67.08 (0.12)          \\
RandTopk-0.2      & 3.84                 & 66.15 (0.08)          \\
RandTopk-0.3      & 3.84                 & 65.09 (0.29)          \\
Size reduction    & 4.00                 & 66.12 (0.09)          \\ \bottomrule
\end{tabular}
\end{table}
\begin{table}[h!]
\caption{Test accuracy and compression ratio on DBPedia.}
\label{table:dbpedia}
\centering
\footnotesize
\begin{tabular}{@{}ccc@{}}
\toprule
Method             & Compressed size (\%) & Test accuracy               \\ \midrule
No Compression     & 100                  & 93.11 (0.13)                \\ 
\midrule
Topk               & 0.44                 & 83.04 (0.45)                \\
RandTopk-0.1       & 0.44                 & 84.88 (0.47)                \\
Size reduction     & 0.50                 & 64.80 (1.09)                \\ 
\midrule
Topk               & 0.88                 & 85.49 (0.35)                \\
RandTopk-0.1       & 0.88                 & 88.01 (0.23)                \\
L1-0.002           & 0.93 (0.01)          & 78.68 (0.83)                \\
Size reduction     & 1.00                 & 78.57 (0.53)                \\
L1-0.001           & 1.08 (0.02)          & 81.35 (0.68)                \\
L1-0.0005          & 1.45 (0.02)          & 87.88 (0.17)                \\ 
\midrule
Topk               & 1.97                 & 87.74 (0.35)                \\
RandTopk-0.1       & 1.97                 & 90.50 (0.11)                \\
Size reduction     & 2.00                 & 86.42 (0.26)                \\ 
\midrule
Size reduction     & 3.00                 & 88.38 (0.10)                \\
Topk               & 3.06                 & 90.05 (0.14)                \\
RandTopk-0.1       & 3.06                 & 91.59 (0.50)                \\
L1-0.0002          & 3.08 (0.09)          & 92.16 (0.10)                \\ 
\midrule
L1-0.0001          & 5.31 (0.24)          & 93.11 (0.05)                \\
Quantization-2bit  & 6.25                 & 91.20 (0.20)                \\ 
\bottomrule
\end{tabular}
\end{table}
\begin{table}[h!]
\caption{Test accuracy and compression ratio on Tiny-Imagenet.}
\label{table:tiny-imagenet}
\centering
\footnotesize
\begin{tabular}{@{}ccc@{}}
\toprule
\multicolumn{3}{c}{Use ImageNet pre-trained weights}           \\ \midrule
Method         & Compressed size (\%) & Test accuracy         \\ \midrule
No compression & 100                  & 75.18 (0.29)          \\ 
\midrule
TopK           & 0.21                 & 70.86 (0.07)          \\
RandTopk-0.1   & 0.21                 & 71.09 (0.14)          \\
Size reduction & 0.23                 & 59.23 (0.83)          \\ 
\midrule
TopK           & 0.42                 & 71.79 (0.54)          \\
RandTopk-0.1   & 0.42                 & 72.15 (0.15)          \\
Size reduction & 0.47                 & 66.52 (0.25)          \\ 
\midrule
TopK           & 0.94                 & 72.52 (0.69)          \\
RandTopk-0.1   & 0.94                 & 73.83 (0.20)          \\
Size reduction & 0.94                 & 68.67 (0.16)          \\
L1-0.0001      & 1.24 (0.04)          & 67.82 (1.15)          \\ \midrule
\multicolumn{3}{c}{Train from scratch}                        \\ \midrule
Method         & Compressed size (\%) & Test accuracy         \\ \midrule
No compression & 100                  & 53.11 (0.18)          \\
TopK           & 0.21                 & 48.36 (0.09)          \\
RandTopk-0.1   & 0.21                 & 50.83 (0.81)          \\
Size reduction & 0.23                 & 35.46 (0.98)          \\ 
\midrule
TopK           & 0.42                 & 47.24 (0.09)          \\
RandTopk-0.1   & 0.42                 & 51.75 (0.04)          \\
Size reduction & 0.47                 & 45.66 (0.29)          \\ 
\midrule
TopK           & 0.94                 & 45.50 (0.19)          \\
RandTopk-0.1   & 0.94                 & 51.16 (0.14)          \\
Size reduction & 0.94                 & 48.87 (0.14)          \\ \bottomrule
\end{tabular}
\end{table}

\clearpage
\section{Derivation of \Cref{eq:min-dist}}
\label{app:eq6}
The $k$-dimensional hypersphere $\Vert \mathbf w \Vert_2 = 1$ has a volume of $2\pi ^{k/2}/\Gamma{(k/2)}$, and each $(k-1)$-dimensional ball $B_i$ has a volume of $2\pi^{(k-1)/2}/\Gamma{[(k+1)/2]}$.
Hence, if there are $n$ balls, the upper bound of minimum center distance (the margin between two classes' weights) $d_W$ shall satisfy
\begin{equation}
    n \cdot \pi^{(k-1)/2}/\Gamma{\left(\dfrac{k+1}{2}\right)} (d_W/2)^{k - 1} \approx 2\pi ^{k/2}/\Gamma{\left(\dfrac{k}{2}\right)},
\end{equation}
then we have
\begin{equation}
    \left(\dfrac{d_W}{2}\right)^{k-1} \approx \dfrac{2{\sqrt{\pi}}}{n}\cdot \dfrac{\Gamma{[(k+1)/2]}}{\Gamma{(k/2)}}.
\end{equation}
Notice the fact that $\Gamma(x + 1/2)/\Gamma(x) \approx \sqrt{x}$, we can further obtain
\begin{equation}
\label{eq:dist-sr}
   d_W \approx 2\cdot \left(\dfrac{2}{n}\sqrt{\dfrac{k\pi}{2}}\right)^{1/(k-1)}.
\end{equation}
In the above analysis, $d_W$ is the estimated maximum distance for size reduction, whose feature space is a $k$-dimensional hypersphere.
On the other side, for top-$k$ sparsification, the feature space contains ${d \choose k}$ hyperspheres since there are total ${d \choose k}$ choices to select $k$ dimensions out of $d$ dimensions.
Also, considering the extra communication for indices, we have to reduce the preserved non-zero elements of top-$k$ from $k$ to $\alpha k$, where $\alpha \in (1/2, 1)$.

Hence, center distance $d_W'$ is changed to
\begin{equation}
\label{eq:dist-topk}
    d_W' \approx {d \choose \alpha k} \cdot 2 \cdot \left(\dfrac{2}{n}\sqrt{\dfrac{\alpha k\pi}{2}}\right)^{1/(\alpha k-1)}.
\end{equation} 

Combining \eqref{eq:dist-sr} and $\eqref{eq:dist-topk}$, we have: 
\begin{equation}
\begin{split}
\label{eq:main-inequality}
\dfrac{d_W'}{d_W} & \approx 
    {d \choose \alpha k} 
    \dfrac{(2/n\sqrt{\alpha k\pi/2})^{1/(\alpha k - 1)}}{(2/n\sqrt{k\pi/2})^{1/(k - 1)}}
    \\
    &= {d \choose \alpha k} 
    \left(\dfrac{2}{n}\sqrt{\dfrac{\alpha k\pi}{2}}\right)^{1/(\alpha k-1)} 
    \cdot
    \alpha^{\dfrac{1}{2}\left(\dfrac{1}{\alpha k - 1} - \dfrac{1}{k-1}\right)}.
\end{split}
\end{equation}

Notice that for if $x, y \in (0, 1)$, then $x^y > x$ (immediately from $y \ln x > \ln x$).
Since $1/(\alpha k - 1),1/(\alpha k - 1) - 1/(k-1) \in (0, 1)$, as long as $2/n \sqrt{\alpha k\pi / 2} < 1$, we have
\begin{equation}
    \left(\dfrac{2}{n}\sqrt{\dfrac{\alpha k\pi}{2}}\right)^{1/(\alpha k-1) - 1/(k - 1)} 
    \ge
    \dfrac{2}{n}\sqrt{\dfrac{\alpha k\pi}{2}},
\end{equation}
and also we have $\alpha^{1/(2(k-1)}\ge \alpha$.
Plug these into \eqref{eq:main-inequality}, we obtain
\begin{equation}
    \dfrac{d_W'}{d_W} \gtrsim {d\choose \alpha k} \dfrac{2}{n}\sqrt{\dfrac{\alpha k\pi}{2}} \alpha = {d\choose \alpha k} \cdot \dfrac{\sqrt{2\alpha^3k\pi}}{n},
\end{equation}
which is \Cref{eq:min-dist}, as we use $d_W$ to denote $d_W^\text{(size reduction)}$ and $d_W'$ to denote $d^{\text{(top-$k$)}}$.


\end{document}